\documentclass{article}
\usepackage{ijcai24}

\usepackage{times}
\usepackage{soul}
\usepackage[hidelinks]{hyperref}
\usepackage[utf8]{inputenc}
\usepackage[small]{caption}
\usepackage{graphicx}
\usepackage{amsmath}
\usepackage{amsfonts}
\usepackage{amsthm}
\usepackage{booktabs}
\usepackage{threeparttable}
\usepackage[switch]{lineno}
\usepackage{multirow}
\usepackage{color}

\usepackage{caption}
\usepackage{subcaption}
\usepackage{url}
\usepackage{multirow}
\usepackage{makecell}
\usepackage{verbatim}
\usepackage{float}
\usepackage{bm}
\usepackage{enumitem}
\usepackage{balance}
\usepackage{tabularx}
\usepackage{algorithm}
\usepackage{algorithmic}
\title{State Machine of Thoughts: Leveraging Past Reasoning Trajectories for Enhancing Problem Solving}

\author{
Jia Liu$^1$\and
Jie Shuai$^2$\footnote{This is work done during an internship at Kwai.}\and
Xiyao Li$^1$
\affiliations
$^1$Kuaishou Technology\\
$^2$Hefei University of Technology\\
\emails
liujia08@kuaishou.com,
shuaijie.hfut@gmail.com,
lixiyao@kuaishou.com
}
\begin{document}

\maketitle

\begin{abstract}

Current Large Language Model-based agents reason within an exploration-evaluation framework, navigating problem-solving processes in a tree-like manner. However, these methods often neglect successful reasoning trajectories once a problem is resolved, leading to inefficient use of these trajectories for future analogous problems. To address this inefficiency, we adopt a state machine to record experience derived from previous reasoning trajectories. Within the state machine, states represent decomposed sub-problems, while state transitions reflect the dependencies among sub-problems. The state machine records both successful and failed trajectories. Utilizing the experience from the state machine, our proposed State Machine of Thoughts (SMoT) selects the most optimal sub-solutions and avoids incorrect ones. Our experiments show that SMoT can significantly improve problem-solving abilities in two exploration-intensive problems: the 24-point game and a taxi navigation reinforcement learning game.

\end{abstract}

\section{Introduction}

In recent years, there has been notable progress in large language models (LLMs) that are built upon large Transformer \cite{transformer} architecture and massive unsupervised training data \cite{llama,GPT3}. Moreover, when appropriately fine-tuned and aligned \cite{instructGPT}, these LLMs have showcased remarkable zero-shot or few-shot performance across various downstream tasks involving semantic understanding, language generation, and reasoning \cite{sparks}. 
% The excellent capacity of LLMs has spurred many research topics, including in-context learning, chain-of-thoughts prompting, and instruction tuning etc., all with the aim of unlocking their full potential and further enhancing their problem-solving ability. 

Whilst LLMs have exhibited impressive performance \cite{llama,GPT3,sparks}, they are sometimes directly produce wrong answers when faced with complex problems. A potential resolution comes from chain-of-thoughts (CoT) prompting approach \cite{self-evaluation,Reflexion,autogen}. These solutions harnesses well-crafted prompts to elicit the LLM's reasoning ability to think intermediate steps, leading to the final accurate answers. Specifically, \citeauthor{CoT} first propose CoT method which provides LLMs with intermediate reasoning step examples as prompts. Through emulation of the CoT prompting, LLMs were able to decompose a complex problem into intermediate easier sub-problems. Through solving the easier sub-problems one by one,  CoT can more accurately deriving the final answers.  Furthermore, \citeauthor{ToT} adopt tree-of-thoughts (ToT) prompting method to implement the ``backtracing'' strategy. ToT explore various sub-solutions for the current sub-problems, followed by the evaluation of each reasoning trajectory to determine whether to continue or backtrack. \citeauthor{GoT} employ a graph-based model to implement the ``divide and conquer'' approach in the reasoning process. In the thought graph, a single node, representing a problem, can be decomposed into several similar sub-problems, each of which can be more effectively handled.

%Additionally, \citeauthor{CoT-SC} propose Self-Consistency enhancement based on the CoT, whereby inferring answers aggregated across multiple reasoning paths serves to further augment the reasoning performance. 
% to explore diverse next sub-question for current choices and then pick up the 
% of intermediary processes based on multiple reasoning paths, in order to determine the most prudent next course of action. 

\begin{figure*}
    \centering
    \includegraphics[width=0.9\linewidth]{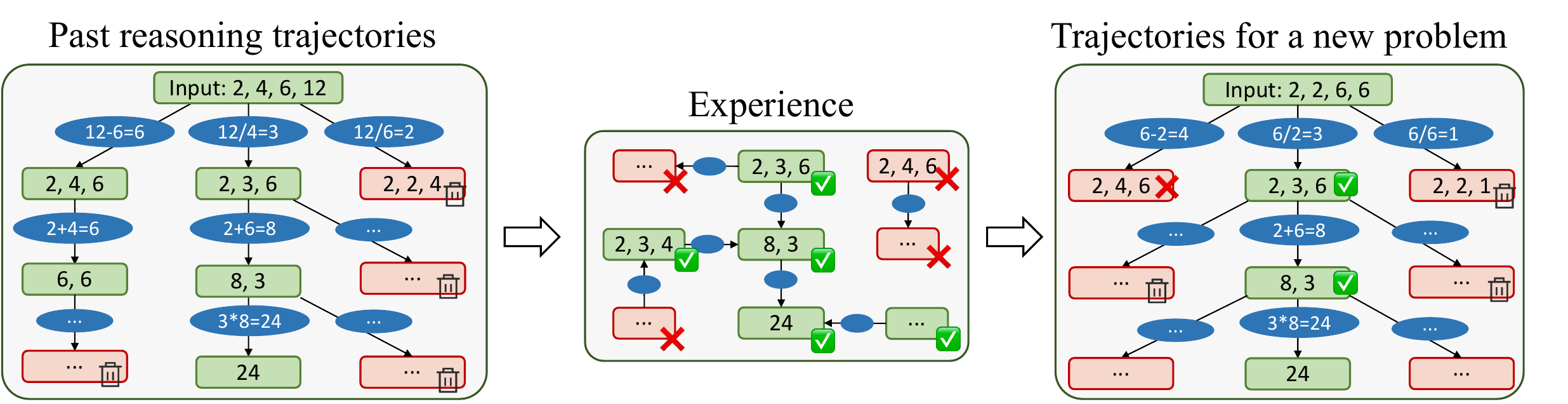}
    \caption{The motivation for improving solving 24-Point Card Games: The goal is to find a way to combine four given numbers to reach the total of 24. By documenting which sets of numbers can successfully achieve or fail to reach this total based on past trajectories, we can store the experience that helps agents quickly decide the best next move when encountering the same numbers in future games. }
    \label{fig:motivation}
\end{figure*}

The ToT prompting method has demonstrated the significance of investigating diverse sub-solutions to effectively tackle problems with constraints, such as the 24-point game. However, despite ToT's effectiveness in enhancing game-solving ability, it often overlooks the utility of experience gleaned from past problem-solving trajectories when addressing similar problems. As depicted in ~\autoref{fig:motivation}, the experience drawn from resolving the array [2, 4, 6, 12] in 24 point game — where subsets like [2, 3, 6] and [3, 8] successfully yield the sum of 24, while a subset such as [2, 4, 6] does not — can be helpful. For instance, when agents facing a new sequence [2, 2, 6, 6], they can deduce that the operation ``6-2=4'' is unlikely to lead to a solution, given that the left numbers [2, 4, 6] do not sum to 24 from previous experience. In contrast, the operation ``6/2=3'' should be selected because it leaves behind the set [2, 3, 6], which can indeed reach the total of 24.
In this paper, we aim to extract and apply insights from previous reasoning trajectories to enhance the problem-solving process. However, this goal is met with two challenges:
\begin{itemize}
    % \item How can we effectively extract and document these experience in a manner that is most synergistic with a LLM? 
    \item How can we extract experience from past trajectories and record experience for best combination with LLM?
    \item In what ways can the extracted experience be applied to improve the reasoning ability for problem-solving? 
\end{itemize}

To this end, we propose State Machine of Thought (SMoT) method to address above two challenges with two steps: (1) To extract experience state machine from past reasoning trajectories; (2) To explore and evaluate with state machine for better problem reasoning. Specifically, we employ a state machine to record experience from prior reasoning paths to handle the first challenge. The state machine \cite{HSM,HSMM}, a well-recognized computational framework, modeling the transitions among various states within the problem-solving process. In our context, we define sub-problems as states, and view reasoning or sub-solutions as the transitions connecting these sub-problems. Specifically, when faced with a sub-problem and its possible transitions to adjacent states, the state machine helps us identify which transition or neighboring state is conducive to resolving the issue and which is not. To extract the experience gained from historical reasoning trajectories, we implement a bidirectional traversal strategy, using both top-to-bottom and bottom-to-top approaches, to distill effective and ineffective trajectories from previous reasoning trees.

% The knowledge gleaned from the state machine confer two benefits: first, they allow us to directly obtain conducive sub-solution and sub-problem without LLM inferring; second, they allow us to eliminate unproductive subsequent sub-problem. Based on these two benefits, we integrate state machine into the two main steps of ToT to prune the exploration tree: sub-solution proposing and sub-problem evaluation. Specifically, at the sub-solution proposing step,  SMoT first search state machine to obtain the next conducive sub-solution. If there exists sub-solutions, SMoT takes the sub-solution and skip LLM proposing. If there is no effective sub-solutions, SMoT follows ToT to propose sub-solution by LLMs. At the sub-problem evaluation step, SMoT will directly remove the unproductive sub-problems which are recorded in the state machine. 

The state machine embedded in SMoT model yields two key advantages: it enables direct retrieval of beneficial sub-solutions and the dismissal of ineffective sub-problems, thus pruning the exploration reasoning tree. These two advantages help us to handle the second challenge. In practice, SMoT consults the state machine to source promising sub-solutions, bypassing LLM inference when possible. If no such sub-solutions are present, SMoT defers to LLMs for generation. Additionally, SMoT uses the state machine to swiftly eliminate known unproductive sub-problems during evaluation. Finally, we compared our proposed SMoT model with state-of-the-art baselines on the 24-point card game and the classic reinforcement learning game of taxi navigation. These two games are conditional constraint problems that require agents to explore as many potential sub-solutions as possible. The experimental results clearly demonstrate SMoT's improvement in problem-solving accuracy and efficiency.

\section{Related Work}

% \subsection{Reasoning with LLMs}

\subsection{Task Decomposition and Planning} 
\citeauthor{CoT} initially discovered that prompting LLMs to decompose a complex problem into subtasks and gradually complete them via chains-of-thought (CoT) could effectively improve LLMs' reasoning performance. Following this line, numerous thought prompting strategies have been proposed to enhance LLMs' reasoning abilities \cite{CoT-SC,ToT,GoT}. \citeauthor{CoT-SC} found that CoT only samples a single reasoning path when answer a question, so the reasoning results suffer from inconsistency issues. Therefore, they proposed the self-consistency strategy, wherein multiple reasoning paths are sampled for a question to boost the reasoning performance of LLMs. 

Although LLMs have exhibited fine planning ability in general domains, some researchers have noted that existing these models still encountering serious difficulties in designing effective plans for domain-specific tasks \cite{zhang2023building,OpenAGI,LLMP}. Therefore, \citeauthor{LLMP} incorporate LLMs to transform the given vision-and-language navigation task into a description in planning domain definition language (PPDL) and then leverages in-domain planners for finding an optimal plan. Similarly, \citeauthor{LLMDP} utilize LLMs to consider high-level plans and incorporate in-domain models to design low-level action sequences. Moreover, a classical path planner are introduced into the robot agent Sayplan \cite{Sayplan} to obtain shorter horizon plan sequences.

\subsection{Reflection and Refinement} 
While existing LLMs possess strong capabilities for task planning, correctly outlining the reasoning paths from the beginning is still challenging. Consequently, researchers have suggested introducing reflection mechanisms whereby the inference process reflects upon and re-plan following paths \cite{selfrefine}. For example, Tree-of-Thoughts proposed by \citeauthor{ToT} structures reasoning paths as a tree. This framework enables LLM agents to self-evaluation and consider diverse reasoning paths during reasoning. \citeauthor{GoT} modelled the reasoning process in graph formulation, introducing three inference operations: aggregation, refining, and generation. Aggregation consolidates several thoughts into an improved one. Refining retrospects and refines the current thought. Generation contemplates subsequent reasoning steps based on the current thought. 

However, only rely on self-evaluation is hard to correctly evaluate current reason path.  Therefore, some works utilize external feedback to evaluation \cite{Voyager,Sayplan,RAP}. For instance, ReAct \cite{ReAct} implemented a three-step reasoning strategy of thought action observe. Here, agents observe the effects of preceding actions upon the environment, think to adjust forthcoming actions, and iterate through this process until arriving at an answer. Reflextion \cite{Reflexion} later introduced a Long-term Memory to keep the outputs from the self-reflection undertaken during this iterative process.

\begin{figure}[!t]
    \centering
    \includegraphics[width=0.95\linewidth]{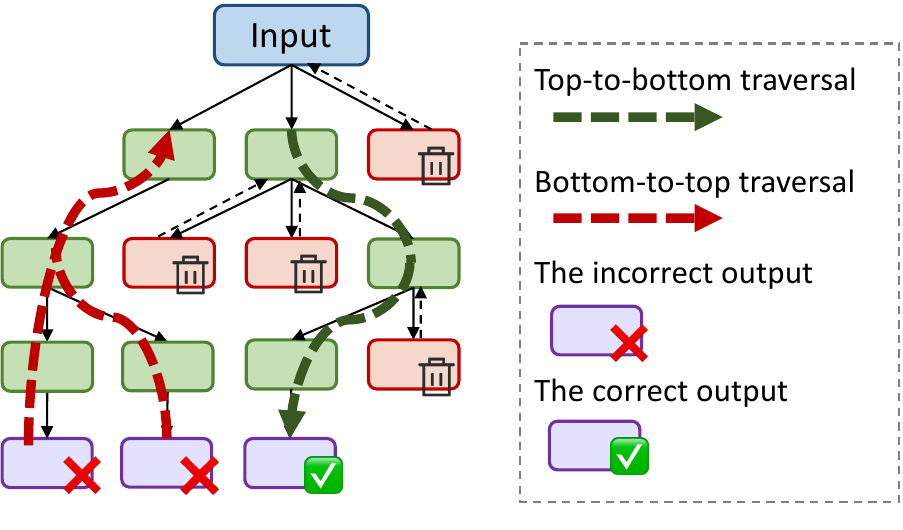}
    \caption{The illustration of two traversal methods for constructing state machine. The green and red box denote the sub-problems evaluated conducive and non-conducive to solve the task, respectively}
    \label{fig:sm_construction}
\end{figure}

\section{State Machine of Thoughts}

In this section, we present a detailed description of our proposed SMoT, which comprises two main components: (1) learning a knowledge state machine from previous reasoning trajectories, and (2) utilizing the prior state machine to reason about new problems.

\subsection{The Construction of the Knowledge State Machine}

Formally, the knowledge state machine is defined as a four-tuple: $\{\mathcal{S}, \mathcal{A}, s^0, \mu\}$, where:

\begin{itemize}
    \item $\mathcal{S}$ is a finite set of symbols denoting sub-problems (i.e., states within the state machine),
    \item $\mathcal{A}$ represents the set of sub-solutions (i.e., actions within the state machine),
    \item $s^0 \in \mathcal{S}$ is the initial problem (i.e., the initial state of the state machine),
    \item $\mu$ is the transition function mapping transitions $\mu: s^{k} \xrightarrow{a^{k+1}} s^{k+1}$, where a sub-problem $s^{k} \in \mathcal{S}$ transitions to another sub-problem $s^{k+1} \in \mathcal{S}$ through the sub-solution $a^{k+1} \in \mathcal{A}$.
\end{itemize}

The state machine records both conducive and non-conducive transitions, i.e., subsequent sub-solutions and sub-problems, given a particular sub-problem. When an agent encounters the same sub-problem again, the state machine can suggest effective sub-solutions and indicate ineffective ones, thus aiding the agent in solving problems both efficiently and accurately.

\subsubsection{Learning the State Machine from Past Reasoning Trajectories}

The Tree-of-Thoughts (ToT) approach adopts a tree structure to explore various sub-solutions while reasoning through complex problems. In the reasoning tree, a complete trajectory from the input node to the output node can be denoted by a transition path $[s^0, (a^1, s^1), \ldots, (a^K, s^K)]$, where state $s^k$ transitions to sub-problem $s^{k+1}$ through the application of sub-solution $a^{k+1}$. As depicted in \autoref{fig:sm_construction}, complete trajectories can be categorized as either successfully solving the problem or failing to do so. Consequently, we propose two methods to utilize these types of trajectories for constructing the knowledge state machine: a top-to-bottom traversal to construct conducive state transitions and a bottom-to-top traversal to construct non-conducive transitions.

\paragraph{Top-to-bottom traversal.}
The top-to-bottom traversal (depicted as the dashed green line in \autoref{fig:sm_construction}) represents the path from the input node to the correct output node. Since these trajectories lead to the correct answers, all transitions within these paths are conducive to problem-solving. Therefore, we record all transitions along these paths in the state machine and denote them with a plus sign $+$ to signify that the states and transitions are conducive, with a typical transition denoted by $s^{k}_+ \xrightarrow{a^{k+1}_+} s^{k+1}_+$.

\paragraph{Bottom-to-top traversal.}
Conversely, the bottom-to-top traversal (illustrated by the red dashed lines in \autoref{fig:sm_construction}) begins from a failed output at the bottom and progresses upwards to the top input node to identify non-conducive transitions. An intermediate node in the tree is deemed non-conducive if all its child nodes are also non-conducive. We implement a "breadth-first" approach starting from the bottom failed output nodes and moving up to the top root nodes. Upon identifying the entire non-conducive trajectories, we document all transitions within these paths in the state machine, marking the states and transitions with a minus sign $-$ as $s^{k}_- \xrightarrow{a^{k+1}_-} s^{k+1}_-$. It is important to note that non-conducive transitions identified from ToT's reasoning trees may not be definitively incorrect, as the sub-solutions proposed for a sub-problem might not encompass all possible solutions. Consequently, conducive trajectories could be misclassified. To mitigate this issue, one potential solution is to sample as many sub-solutions as possible.

\paragraph{The state query function.}
To equip LLM-based agents with insights from the state machine, we introduce two state query functions:

\begin{itemize}
    \item \textbf{Conducive Sub-solution Query}: Function \( f_{s}(s^k) \) takes a sub-problem \( s^k \) as input and produces a sequence of tuples \( \left[(a^{k+1}_1, s^{k+1}_1), \ldots, (a^{k+1}_N, s^{k+1}_N)\right] \), where each tuple contains a next sub-solution \( a^{k+1}_i \) and the subsequent sub-problem \( s^{k+1}_i \), for \( i = 1, \ldots, N \). This function aims to offer sub-solutions that have been empirically proven to be conducive to solving the problem in past reasoning trajectories.
    \item \textbf{Sub-problem Solvability Query}: Function \( f_{p}(s^k) \) accepts a sub-problem \( s^k \) and returns \( 1 \) if the sub-problem is conducive to solving the overall problem, and \( 0 \) otherwise. This binary indicator quickly informs the agent about the potential worth of pursuing the sub-problem within the context of the broader problem-solving effort.
\end{itemize}

These query functions act as interfaces between the agent's current context and the accumulated knowledge within the state machine, enabling the agent to make informed decisions based on historical data.

\subsection{Reasoning with the Prior State Machine}

Existing prompting methods, such as ToT, adopt a paradigm of proposing sub-solutions and evaluating sub-problems to identify feasible reasoning trajectories. By constructing a state machine from past trajectories, we can enhance agents' capabilities in two ways: (1) conducive sub-solutions and sub-problems can directly offer optimal sub-solutions for the agent; (2) non-conducive sub-solutions can be accurately evaluated as ineffective.

\paragraph{State machine enhanced sub-solution proposer.}
The sub-solution proposer $\mathcal{P}(s^{k})$ aims to propose sub-solutions and the corresponding transited sub-problems $\big[({a}_{n=1}^{k+1}, s_{n=1}^{k+1}), \ldots, ({a}_{n=N}^{k+1}, s_{n=N}^{k+1})\big]$, given the current sub-problem $s^{k}$. In this step, we utilize the conducive sub-solution query function $f_{s}(s^{k})$ to query the state machine for optimal sub-solutions. If sub-solutions exist in the state machine, SMoT directly adopts these solutions instead of generating new ones using LLMs. Conversely, if no sub-solution is present in the state machine, SMoT reverts to the ToT approach, which involves proposing sub-solutions using LLMs $\mathcal{P}_l(s^{k}, p_{\theta})$, where $p_{\theta}$ denotes the LLM with parameter $\theta$.

\paragraph{State machine enhanced next sub-problem evaluator.}
Upon obtaining potential sub-solutions and their corresponding transited new sub-problems, the sub-problem evaluator $\mathcal{V}(s^{k}) \rightarrow v$ is adopt to evaluate sub-problems, where $v$ denotes the sub-problem's solvability. In SMoT, the evaluator provides four levels of solvability scores:
\begin{itemize}
    \item \textbf{Absolutely solvable}: This score indicates that the sub-solution proposed from the state machine is assuredly solvable and requires no further evaluation.
    \item \textbf{Possible}: This score, evaluated by LLMs, indicates that the sub-problem, not recorded in the state machine, has possible solvability.
    \item \textbf{Impossible}: This score, evaluated by LLMs, indicates that the sub-problem, not recorded in the state machine, is unsolvable.
    \item \textbf{Absolutely unsolvable}: This score indicates that the sub-solution is recorded as unproductive in the state machine.
\end{itemize}
SMoT's evaluator is compatible with the existing ToT method. Specifically, if the sub-problem's solvability has been recorded in the state machine (checked by the sub-problem solvability query function $f_{p}(s^k)$), we rely on the solvability recorded in the state machine. If not, the evaluator reverts to the ToT evaluator $\mathcal{V}_l(s^{k}, p_{\theta}) \rightarrow v$, where $p_{\theta}$ is the LLM with parameter $\theta$.

\subsubsection{The Search Algorithm of SMoT}

SMoT is compatible with the existing ToT method, allowing both breadth-first search (BFS) and depth-first search (DFS) algorithms to be applicable for SMoT. Due to space constraints, we only introduce the BFS implementation of SMoT in Algorithm \ref{alg:algorithm} .

% \begin{algorithm}
% \caption{SMoT-BFS search algorithm }
% \begin{algorithmic}

% \Require $n \geq 0$
% \State $y \gets 1$
% \State $X \gets x$
% \State $N \gets n$

% \end{algorithmic}
    
% \end{algorithm}

\begin{algorithm}[tb]
    \caption{SMoT-BFS search algorithm}
    \label{alg:algorithm}
    \textbf{Input}: the raw problem $s^0$, conducive sub-solution query function $f_s$,  sub-problem solvability query function $f_p$, LLM-based proposer $\mathcal{P}_{l}$, LLM-based evaluator $\mathcal{V}_l$, LLM $p_{\theta}$, step limit $K$, and breadth limit $B$ 
    % \textbf{Output}: Your algorithm's output
    \begin{algorithmic}[1] %[1] enables line numbers
        \STATE Let $\mathcal{Q} \xleftarrow{} \{s^0\}$.
        % \FOR{$k=1, \dots, K$}A \cup \{x\}
        \STATE \textbf{for} $k=1, \dots, K$ \textbf{do}
            % \STATE 
            \STATE \quad /* sub-solution proposing */
            \STATE \quad \textbf{for} $s^k_n \in Q$ \textbf{do}
                \STATE {\quad \quad \textbf{if} $f_s(s^k_n)$ is not empty \textbf{then}}
                    \STATE\quad \quad \quad $Q'^k \cup f_s({s^k_n})$ 
                \STATE \quad \quad \textbf{else}
                    \STATE \quad \quad \quad  $Q'^k \cup \mathcal{P}_l(s^k_n, p_{\theta}, B)$  
            \STATE \quad /* sub-problem evaluation $\mathcal{V}(s^k_{n}) \xrightarrow{} v$ */
            \STATE \quad \textbf{for} $s^k_{n} \in Q'^k$ \textbf{do}
                \STATE {\quad \quad \textbf{if} $s^k_{n}$ in $f_p$ \textbf{then}}
                    \STATE \quad \quad \quad $v^k_{n} \xleftarrow{} f_p({s^k_{n}})$ 
                \STATE \quad \quad \textbf{else}
                    \STATE \quad \quad \quad  $v^k_{n} \xleftarrow{} \mathcal{V}_l(s^k_{n}, p_{\theta})$ 
            \STATE \quad /* choose most conducive sub-problems */
            \STATE \quad $Q^k \xleftarrow{} \operatorname{argmax}_{s^k_{n} \in Q'^k} \mathcal{V}(s^k_{n})$
        % \ENDFOR
        \STATE \textbf{return} solution
    \end{algorithmic}
\end{algorithm}

\begin{figure*}[ht!]
    \centering
    \includegraphics[width=0.85\linewidth]{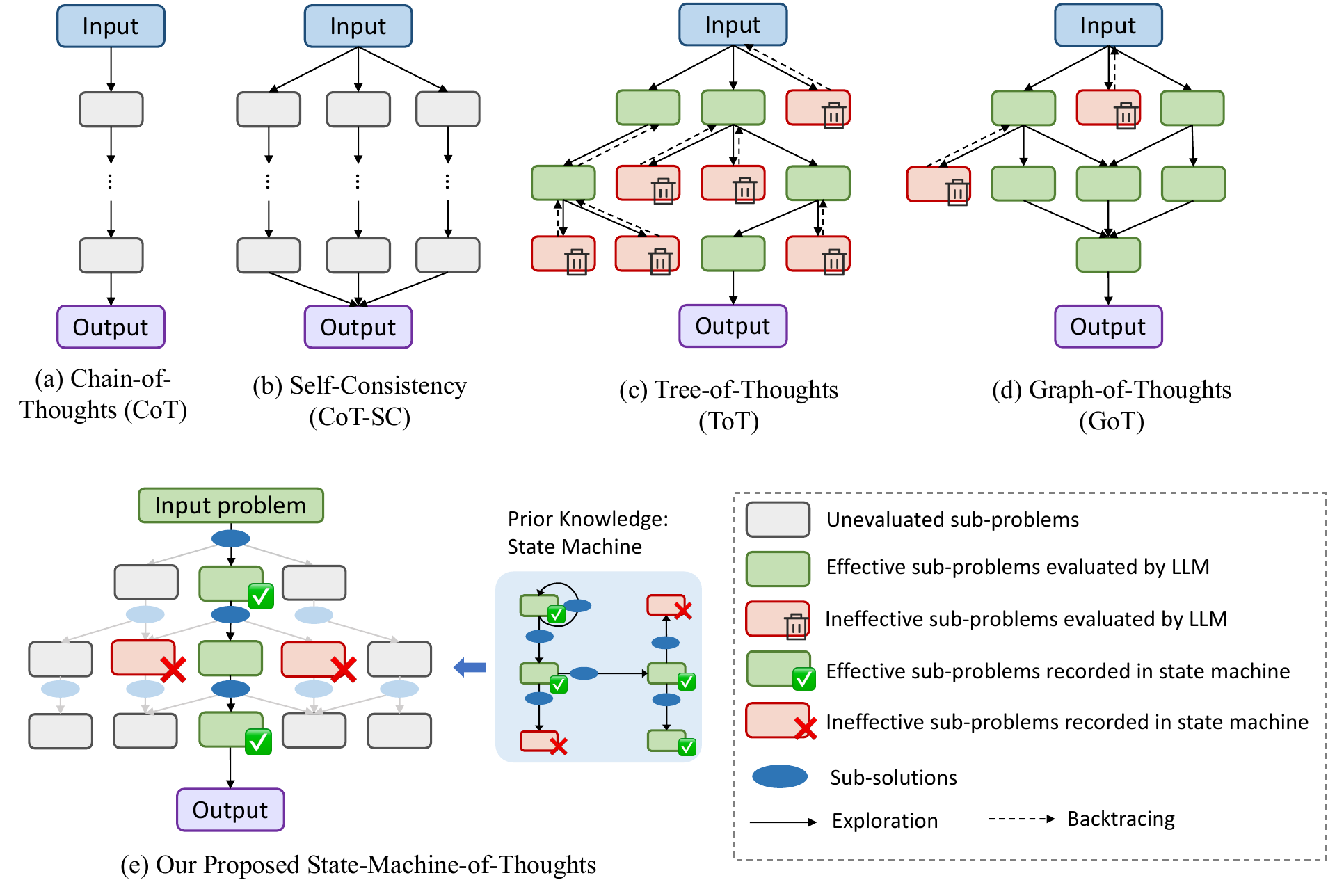}
    % \vspace{-4mm}
    \caption{Comparison of State Machine of Thoughts (SMoT) with other prompting strategy}
    \label{fig:prompt-comparison}
\end{figure*}

\section{Comparison between SMoT and Existing Prompting Approaches}

In this section, we elaborate on the conceptual framework of SMoT and compare its features with those of existing prompting methods in ~\autoref{fig:prompt-comparison}.

The Chain-of-Thought (CoT) approach \cite{CoT} provides Large Language Models (LLMs) with a series of few-shot examples that delineate a step-by-step reasoning process. The LLMs are then expected to emulate a similar reasoning sequence when addressing a query. Self-Consistency (CoT-SC) \cite{CoT-SC} seeks to enhance the basic greedy decoding strategy utilized in CoT prompting by sampling multiple diverse reasoning trajectories through few-shot CoT and electing the most consistent response from the generated answers.

Tree of Thoughts (ToT) \cite{ToT} employs a method of exploration, evaluation, and backtracking as it searches for the appropriate reasoning pathway. Should the current trajectory prove insufficient in conclusively determining the answer, it will backtrack to a previous step or even earlier in the process to explore alternative viable routes toward an optimal solution. This technique is particularly effective for problems that require constraint satisfaction, such as the 24-game or crossword puzzles. Nevertheless, for issues where the reasoning pathway is pre-established, the extra steps of exploration and backtracking not only contribute to unnecessary computational effort but also amplify the uncertainty within the reasoning process.

Graph of Thoughts (GoT) \cite{GoT} adopts a divide-and-conquer strategy, breaking a multifaceted problem into a series of simpler sub-problems until these sub-problems can be effortlessly resolved. Consequently, the GoT methodology is especially apt for problems that can be divided, including tasks like sorting, set intersection, and keyword counting. The decomposition of the central problem simplifies the resolution process by allowing for the recursive solution of its independent subcomponents.

In contrast to ToT or GoT, which require supplementary reasoning steps for exploration and backtracking, SMoT capitalizes on a state machine as prior knowledge for problem-solving. The state machine constructs a bridge that applies experience from past reasoning trajectories to assist LLMs in reasoning subsequent similar problems. This structured guidance ensures that LLM reasoning remains within predefined exploration boundaries, thus enabling the avoidance of unproductive sub-problems at an early stage and the selection of sub-solutions that have been deemed conducive to resolving the task.

\section{Experiments}

In this section, we compare SMoT with other existing prompting methods on two tasks that pose significant challenges to current Large Language Models (LLMs): a toy reinforcement learning game named ``Taxi Navigation'' and a numerical reasoning game called ``24-Point Card Gam''. Both games require diverse exploratory strategies during the intermediate reasoning steps, and the same intermediate sub-problems often recur across different tasks. Unless specified otherwise, our experiments were conducted using the GPT-3.5-turbo LLM with a sampling temperature set to 0.7.

% The primary challenge within these problems stems from the presence of walls on the map, which prevent direct access between any two points, necessitating that the LLMs learn to navigate longer paths to circumvent these obstacles. 

\subsection{Taxi}

The Taxi environment is a classical reinforcement learning problem\footnote{\url{https://gymnasium.farama.org/environments/toy_text/taxi/}} that involves navigating a taxi to passengers in a $5 \times 5$ grid world, picking them up, and dropping them off at one of four designated colored locations. As depicted in \autoref{fig:taxi_settings}(a), the passenger is initially at one of the colored locations, and their destination is another colored location. The environment provides observations that include the taxi's coordinates, the passenger's location, and their destination. The action space encompasses four movement directions (north, south, east, and west) and actions to pick up or drop off passengers. The episode concludes once the passenger is dropped off at their destination. The success of the mission depends upon delivering the passenger to their destination. 

\subsubsection{Task Setup} 

\begin{figure}
    \centering
    \includegraphics[width=\linewidth]{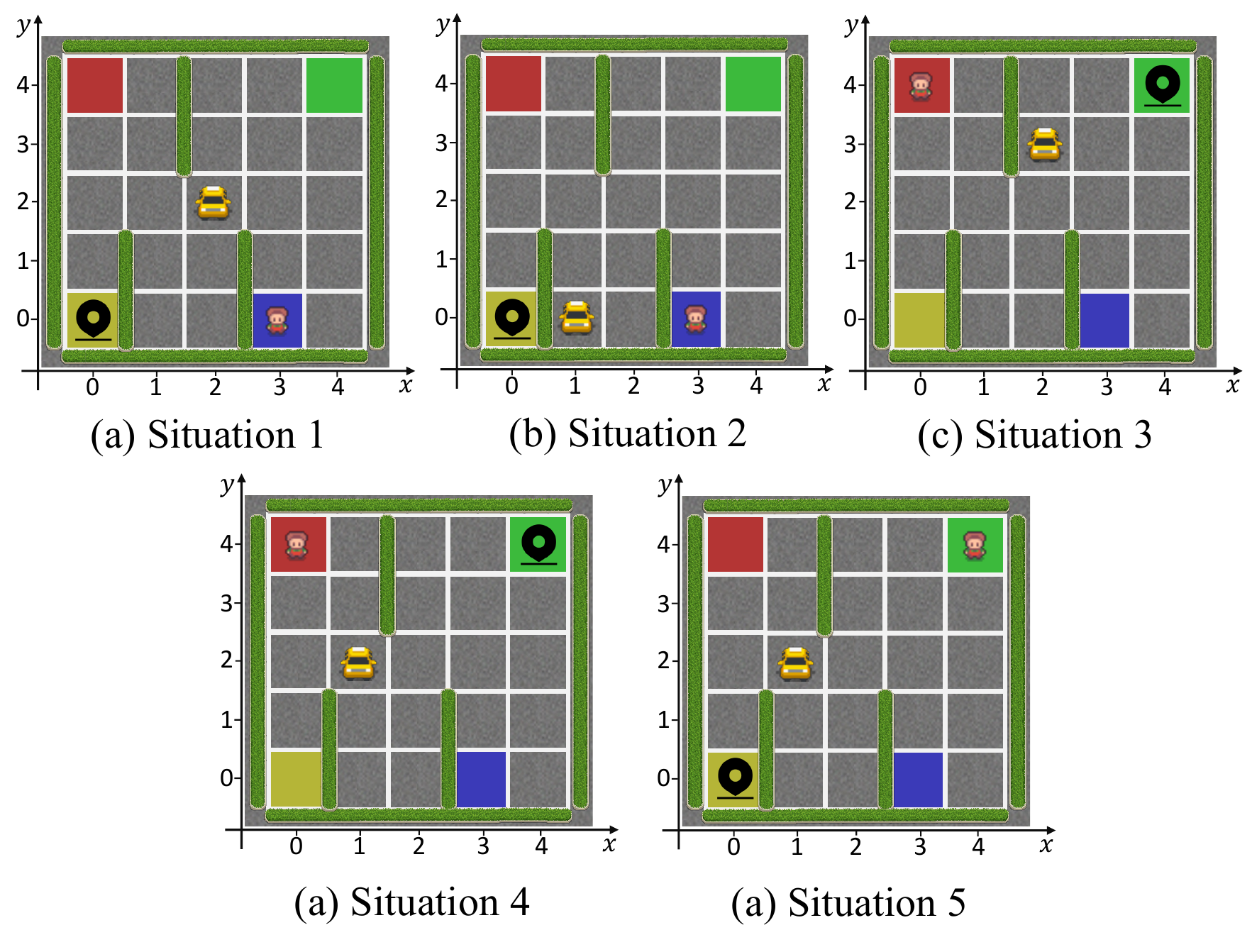}
    % \vspace{-4mm}
    \caption{Five manually designed situations for experiments}
    \label{fig:taxi_settings}
\end{figure}

To ensure a fair comparison, we establish five challenging scenarios, illustrated in Figure \ref{fig:taxi_settings}, where the taxi's starting position, the passenger's origin, and the desired destination are deliberately predetermined. For each scenario, we evaluate a method 20 times. We utilize the task success ratio and the number of LLM inferences to assess accuracy and efficiency.

\subsubsection{Baselines}
% For the taxi navigation task, we selected \textbf{CoT} and \textbf{ToT} as comparative methods, where ToT is implemented by BFS (Breadth-first search). Standard, CoT, and CoT-SC demonstrated poor performance during our preliminary exploration and were therefore excluded from consideration. At each step, LLM is prompt with the observations from the environment. Another notable method is RaAct ~\cite{ReAct}, which reasons through a three step process of thinking, action, and observation. However, through empirical testing we found that RaAct's thinking step struggles to adequately analyze previous routes, sometimes resulting in the taxi driving in endless circles.
For the taxi navigation task, we selected \textbf{CoT} (Chain-of-Thought) and \textbf{ToT} (Tree-of-Thought) as comparative methods, where ToT is implemented using Breadth-first Search (BFS). Standard CoT, as well as CoT-SC, exhibited suboptimal performance in our preliminary trials and were hence excluded from further analysis. At each step, the LLM is prompted with observations from the environment. Another method of note is RaAct\cite{ReAct}, which employs a three-step reasoning process involving thought, action, and observation. However, empirical tests revealed that RaAct's thinking phase often fails to effectively consider previous routes, occasionally leading to the taxi driving in loops.

\subsubsection{SMoT Implementation Details}

\paragraph{The construction of knowledge state machine.}
Initially, we utilize ToT to explore the effective trajectories, which describe the paths from all coordinates to all colored locations, such as, [(2,2), (west, (1,2)), (north, (1,3)), ... , (west, red location)]. After gathering these valid trajectories, we utilize the top-to-bottom traversal method to construct the state machine which only contains the conducive move for each coordinate to every colored locations. The number of states recorded in state machine is 96. 
\paragraph{Construction of the knowledge state machine}
Initially, we employ ToT to explore effective trajectories that describe paths from all coordinates to all colored locations, such as {[(2,2), (west, (1,2)), (north, (1,3)), ..., (west, red location)]}. After compiling these valid trajectories, we use a top-to-bottom traversal method to construct a state machine that includes only the beneficial moves for each coordinate to every colored location. The state machine records a total of 96 states.

\paragraph{Reason with the prior state machine.}

During each step of exploration, the LLM is prompted with the current observations from the environment, which encompass the taxi's coordinates, the passenger's location, and their destination. To leverage the knowledge state machine, the LLM is also provided with advantageous movements from the taxi's current location to the four colored locations.

\begin{table} \small
\caption{The performance comparison of the taxi task in terms of success ratio and the average count of LLM infers}
\begin{tabular}{@{}lrrrrrr@{}}  
\toprule
\multirow{2}{*}{Methods} & \multicolumn{3}{c}{Success ratio}                                            & \multicolumn{3}{c}{\# LLM infers}                                          \\ \cmidrule(l){2-7} 
                         & \multicolumn{1}{r}{CoT} & \multicolumn{1}{r}{ToT} & \multicolumn{1}{r}{SMoT} & \multicolumn{1}{r}{CoT} & \multicolumn{1}{r}{ToT} & \multicolumn{1}{r}{SMoT} \\ \midrule
Situation 1              & 0\%                     & 20\%                    & \textbf{100\%}                    & 30.0                      & 25.6                      & \textbf{12.0}                     \\
Situation 2              & 0\%                     & 50\%                    & \textbf{95\%}                     & 30.0                      & 39.5                        & \textbf{15.2}                     \\
Situation 3              & 0\%                     & 35\%                   & \textbf{100\%}                    & 30.0                      & 56.2                         & \textbf{15.0}                     \\
Situation 4              & 0\%                     & 30\%                    & \textbf{100\%}                    & 30.0                      & 33.6                        & \textbf{15.0}                     \\
Situation 5              & 0\%                     & 15\%                      & \textbf{100\%}                    & 30.0                      & 30.9                        & \textbf{15.0}                     \\
Average                  & 0\%                     & 30\%                      & \textbf{99\%}                     & 30.0                      & 37.2                       & \textbf{14.4}                     \\ \bottomrule
\end{tabular}
\label{tab:taxi_performance}
\end{table}

\subsubsection{Overall Performance} 
Table~\ref{tab:taxi_performance} presents a comparison of overall performance. We observe the following: Firstly, the CoT method was rarely able to complete the task successfully, with the number of LLM inferences consistently reaching the preset maximum limit of 30. 
Secondly, the ToT method exhibited a significant improvement with a 30\% success ratio, suggesting that the exploration of different actions is critical for this task.
Thirdly, the SMoT method achieved the best performance in terms of both accuracy and efficiency. 
This experiment demonstrates that for tasks with a high repetition rate of sub-problems, SMoT can achieve accurate and efficient reasoning for the task by preliminarily exploring all sub-problems and accumulating experience.
% This experiment demonstrates that incorporating a state machine equipped with prior knowledge can markedly enhance the accuracy and efficiency of LLM-based inferences for specific tasks.

\subsection{24-point Card Games}

The 24-Point Card Game is a popular mathematical card game that challenges players to manipulate four given numbers. The objective is to combine these numbers using the arithmetic operations of addition, subtraction, multiplication, and division to achieve a total of 24. Each number must be used exactly once. This task has been adopted to test the numerical reasoning abilities of agents in ToT method \cite{ToT}.

\subsubsection{Task Setup}

We follow the task setup in ToT \cite{ToT}, where the experiments on conducted on the relatively hard games indexed 901-1,000 problems from the \url{4num.com}. For each problem, the output equation is equal to 24 will be regard as success otherwise be fail.

\subsubsection{Baselines}

Following the baseline selection in the taxi problem and the experiments in ToT, we adopt CoT and ToT as our baselines. CoT consists of prompts with several task reasoning trajectories. The implementation of ToT involves two main steps: sub-solution proposal and sub-problem evaluation. The sub-solution step involves selecting two numbers and a mathematical operation to calculate a new number, with a breadth limit $B$ set to 20. The evaluation step involves judging the solvability of the sub-problem, which is the possibility of the given number reaching 24. The LLM adopted in this paper is GPT-3.5-turbo, which exhibits relatively weak numerical reasoning capabilities. Therefore, we employ Python to infer the equation from a given ToT reasoning trajectory.

\subsubsection{SMoT Implementation Details} 

\paragraph{Construction of the knowledge state machine.}
We utilize the baseline tool ToT to explore trajectories that describe the paths from the input four numbers to the final answer. For example, a trajectory might look like [(1,2,3,4), (1*2=2, (2, 3, 4)), (2*3=6, (6, 4)), (6*4=24, (24))] where each mathematical operation is considered a sub-solution and the remaining numbers are the subsequent sub-problems. Owing to the time efficiency required for the ChatGPT API, we collected these trajectories from the indexed problems 1-900, using a Python script with a backtracking algorithm. After collecting these trajectories, we employed both top-down and bottom-up traversal algorithms to construct state machines, which include both beneficial and non-beneficial sub-solutions. In our experiments, the number of sub-problems recorded in the state machine reached 15,310.

\paragraph{Reasoning with the prior state machine.}
At the sub-solution proposing stage, SMoT initially adopts sub-solutions from the state machine without subsequent evaluation. If the state machine lacks sub-solutions, SMoT reverts to ToT and uses an LLM to propose sub-solutions. During the sub-problem evaluation stage, SMoT discards non-beneficial sub-solutions and sub-problems recorded in the state machine.

\begin{table}[t]  \small
\centering
\caption{The overall performance of 24-point game}
\begin{tabular}{@{}lrr@{}}
\toprule
     & Success Ratio & \#LLM infers \\ \midrule
CoT  & 0\%           & 3.0            \\
ToT  & 20\%          & 88.8        \\
SMoT & \textbf{56\%}          & \textbf{36.2}        \\ \bottomrule
\end{tabular}
\label{tab:24_overall}
\end{table}

\subsubsection{Overall Performance}

As illustrated in Table \ref{tab:24_overall}, the performance on this task was extremely poor, with a 0\% success rate, attributed to a lack of exploration of the diverse intermediate steps. The ToT methodology demonstrated a marked improvement over CoT, in line with the experimental trends observed within ToT. However, there was a notable increase in the number of LLM reasoning times required by ToT. Our proposed SMoT approach achieved the best success ratio of 56\%, and the average number of LLM reasoning iterations was significantly reduced to 36.2, considerably less than that required by ToT. This emphatically underscores the superiority of incorporating prior SM knowledge in enhancing the success rate of problem-solving while simultaneously reducing the number of LLM reasoning iterations needed.

\begin{table}[t]  \small
\centering
\caption{The ablation study of SMoT with different state numbers}
\begin{tabular}{lrr}
\hline
                           & Success Ratio & \#LLM infers \\ \hline
SMoT-(100\% states) & 56\%          & 36.2        \\
SMoT-(80\% states)  & 55\%          & 40.5        \\
SMoT-(60\% states)  & 56\%          & 41.6        \\
SMoT-(40\% states)  & 57\%          & 50.5        \\
SMoT-(20\% states)  & 50\%          & 60.1        \\
SMoT-(10\% states)  & 52\%          & 67.8        \\
SMoT-(5\% states)   & 42\%          & 72.2        \\
SMoT-(1\% states)   & 30\%          & 86.3        \\
ToT                        & 20\%          & 88.8        \\ \hline
\end{tabular}

\end{table}

\subsubsection{Ablation Study of SMoT with Varying Numbers of States}

To further investigate the impact of the number of states in a state machine on SMoT's problem-solving capabilities, a series of experiments were conducted on the game of 24-point using different numbers of states. The setups ranged from 100\% to 1\% of the original state count (as shown in Table 3). The data indicates an upward trend in the average number of LLM inferences as the number of states decreases. Regarding the success rate, SMoT maintained a relatively steady performance, around 56\%, when the state count was reduced from 100\% to 40\%. However, a noticeable decline in the success rate begins when the state count falls to 20\% and below, dropping from 50\% to 30\%. Notably, at 1\% state retention, SMoT achieved a 30\% accuracy rate, significantly outperforming the ToT model's 20\% accuracy. These findings suggest that SMoT retains a considerable problem-solving capacity even with a severely limited number of states, outpacing the baseline ToT model in performance.

\begin{table}[t]  
\centering
\caption{The robustness study of SMoT with noisy state machine}
\begin{tabular}{lrr}
\hline
                           & success ratio & \#LLM infers \\ \hline
% SMoT-(100\% state machine) & 56\%          & 36.27        \\
SMoT-(80\% noise)  & 30\%          & 51.7        \\
SMoT-(60\% noise)  & 41\%          & 47.0        \\
SMoT-(40\% noise)  & 38\%          & 55.7        \\
SMoT-(20\% noise)  & 43\%          & 39.0        \\
SMoT-(0\%  noise)  & 56\%          & 36.2        \\ \hline
\end{tabular}
\end{table}

\subsubsection{The Robustness Study for Noisy Knowledge State Machine}

% To verify the robustness of SMoT integrating with noisy experience state machine. 我们将state中的 conducive sub-solution and sub-problems 按照比例改为 non-conducive. 表4 reports SMoT 在 20\% 至 80\% noise条件下的表现。可以看到，引入噪声后，SMOT的成功率有了明显下降，但是任然高于ToT的成功率。此外，SMoT在20\%,40\%和 60\%比例噪声的情况下，成功率维持在40\%左右。当噪声比例高达80\%时，成功率降低到了30\%。 
To evaluate the robustness of SMoT when incorporating noisy experience in the state machine, we transformed a proportional amount of conducive sub-solutions and sub-problems into non-conducive ones. Table 4 presents the performance of SMoT in conditions with noise levels ranging from 20\% to 80\%. It is observed that the introduction of noise significantly reduces SMoT's success rate, yet it remains superior to that of ToT baseline. Furthermore, SMoT sustains a success rate around 40\% amidst 20\%, 40\%, and 60\% noise levels. At a high noise threshold of 80\%, the success rate falls to 30\%, underscoring the adverse effects of incorrect state machine on the efficacy of SMoT.

\section{Conclusion}
% \subsection{Limitations and future directions}

% Our proposed SMoT improve Tree-of-Thoughts method by integrating the experience extracted from previous reasoning trajectories. However, SMoT employs state machine to record experience, making it less applicable to situations that the number of sub-problem is really huge that is hard to save. 
% However, in manly in-domain situations or real-world contexts, state machines has been widely employed, such as in robotics and business process management. 
% Deploying LLMs in these domains inevitably involves modeling complex state transitions, thus presenting new opportunities and challenges for the combine of LLMs with state machines. 

% \blueText{On the other hand, SMoT which heavily rely on state machine naturally fit to analysis and handle the sequence issues. Therefore, the reasoning process cannot be efficiently partitioned and handled in parallel. }

% Moreover, existing SMoT solutions rely on manual design of state machines to reason about unfamiliar problems. Designing state machines suitable for LMs requires continual debugging and adjustment. Consequently, how to leverage the reasoning ability of LLM (Large Language Model) to assist in the design of state machines becomes a pivotal research question. 

% \subsection{Future Directions}
% 强调SMoT在针对具有状态转移任务中的应用。特别是在工业界。

% \subsection{Conclusion}

In this paper, we introduced the State Machine of Thought (SMoT) paradigm, which leverages  experience extracted from past reasoning trajectories to guide LLMs in effective problem-solving.
Specifically, we utilize state machine to record the experience, where states represent the decomposed sub-problems and state transitions denote the sub-solutions. Integrate with the experience state machine, SMoT could will well handle the same sub-problems which are recorded in the state machine. The experiment conducted on a toy reinforce learning game and 24-point numerical reasoning task clearly demonstrate the effectiveness and efficiency of proposed SMoT. 
% The major contribution of our paper is that we open up new direction in utilizing the past reasoning trajectories for better 
% Through experiments on an array reasoning task and a classical reinforcement learning task, we demonstrated that SMoT outperforms the state-of-the-art baseline method, achieving a remarkable improvements in terms of accuracy and efficiency. These results highlight the effectiveness of leveraging pre-existing knowledge and the SMoT paradigm in enhancing LLMs' problem-solving capabilities.
% Our work opens up new direction for further research in utilizing pre-established methods to guide LLM reasoning within predefined boundaries, ultimately unlocking the full potential of LLM agent.

%% The file named.bst is a bibliography style file for BibTeX 0.99c
\bibliographystyle{named}
\bibliography{ijcai24}

\end{document}